\RequirePackage{placeins}
\RequirePackage{amsmath}
\RequirePackage{pifont}
\AtEndDocument{%
  \raggedbottom
  \begin{figure}[H]
    \centering
    \includegraphics[width=\linewidth]{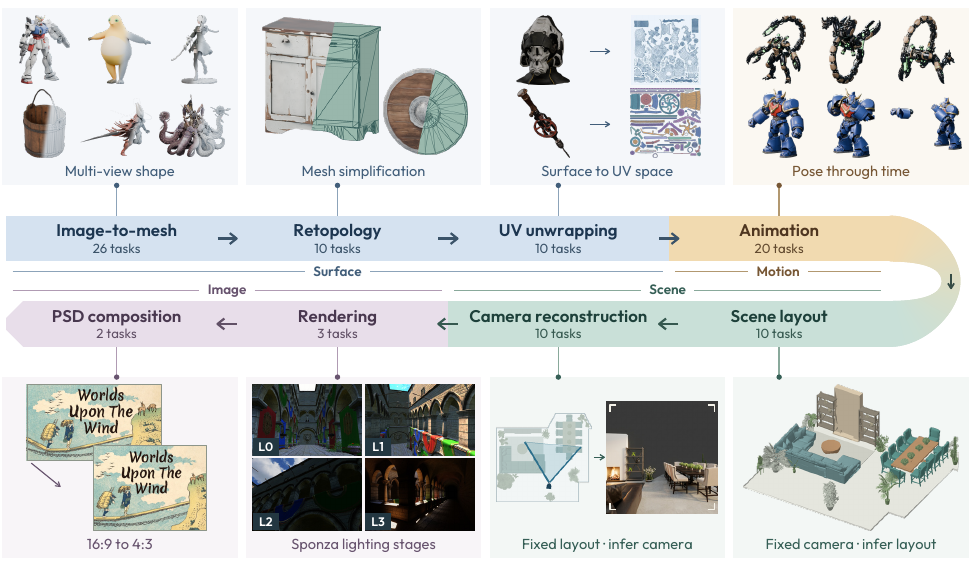}
    \caption{GameAtlas evaluates visual coding and spatial reasoning across eight disciplines (91 tasks). The layout and camera examples share a living and dining room. L0--L3 show the four Sponza rendering stages.}
    \label{fig:gameatlas-overview}
  \end{figure}
  \clearpage
  \setlength{\parskip}{4pt plus 1pt minus 1pt}%
  \setlength{\intextsep}{9pt plus 2pt minus 2pt}%
  \setlength{\textfloatsep}{11pt plus 2pt minus 2pt}%
  \setlength{\bibsep}{2pt plus 1pt}%
  \pagestyle{plain}\section{Introduction}

Evaluating a coding agent on a visual task requires asking how it works from observation to implementation. Multiple views show how an object appears from different directions, leaving the agent to infer a single 3D structure consistent with them~\citep{spatialbench}. A motion sequence shows how a body changes pose, leaving the agent to infer both how its parts move and which contacts they maintain. Inferring these relationships calls for spatial reasoning, while realizing them in geometry and motion calls for visual coding. The precision of both becomes apparent in details that an otherwise plausible result can conceal. A model may match a frontal silhouette yet have the wrong depth in profile~\citep{eval3d}, just as an animation may reproduce the key poses of a stride while the planted foot slides between them~\citep{kimodo,ground}. How precisely, then, does an agent's code reproduce the geometric, spatial, and temporal relationships implied by a visual reference?

Game development places substantial demands on these capabilities. Each frame depends on decisions made across modeling, layout, animation, and camera placement. End-to-end benchmarks evaluate the combined effect of those decisions by running games built from specifications and replaying the intended interactions~\citep{gamecraft}. A finished game, however, offers limited evidence about the accuracy of any individual judgment. An object may appear too small because of its modeled dimensions, its placement in the scene, or the camera's field of view. One-shot generation also yields a different kind of deliverable from the one a professional production requires. A studio framing a shot in an existing scene needs an editable camera setup that matches the intended composition with the approved geometry and layout in place. That setup must support further adjustments as the shot goes through review. Delivering a playable game in a single pass does not fulfill that brief. The distinct stages of professional development therefore provide a natural framework for evaluation. Each stage isolates a decision with observable consequences and allows us to test whether an agent can refine its work to meet the requirements of the next stage.

With \textbf{GameAtlas}, we bring the evaluation of coding agents back to the fundamentals of visual coding and spatial reasoning. We decompose eight stages of professional game development into 91 independently verifiable tasks spanning image-to-mesh reconstruction, retopology, UV unwrapping, animation, scene layout, camera reconstruction, rendering, and PSD composition. Each task specifies the assets and references an agent receives, the conditions it must hold fixed, and the artifact it must deliver. Together, these choices isolate the visual or spatial relationships at stake in that task. In professional production, the requirements for a deliverable guide both its refinement and its acceptance. GameAtlas reflects this workflow by allowing agents to inspect their work against the brief, decide what to revise or preserve, and refine their implementation within a fixed submission budget. Game development experts translate the delivery requirements into executable checks on geometry, motion trajectories, scene parameters, or rendered views, so that artifacts can be evaluated directly without an LLM judge. Examining an agent's revisions alongside its final delivery lets us assess both the precision of its work and its ability to use visual evidence and available feedback to correct errors.

Our contributions are fourfold.
\begingroup
\setlength{\leftmargini}{16pt}
\setlength{\labelsep}{8pt}
\begin{enumerate}
\setlength{\itemsep}{0.18em}
\item We introduce GameAtlas to evaluate how accurately coding agents infer geometric, spatial, and temporal relationships from visual references and realize them through code. Its 91 tasks span eight stages of professional game development. By fixing some elements of a production problem and requiring the agent to recover others, each task isolates a specific set of relationships for evaluation.
\item Game development experts define the relationships that a correct deliverable must preserve and encode them as executable checks on the artifact itself. This yields a fine-grained measure of which constraints an agent's implementation satisfies and where it departs from the reference, without an LLM judge.
\item GameAtlas assesses whether agents can use evidence from their own outputs and the feedback available to them to improve an initial implementation. Within a fixed submission budget, agents decide what to inspect, what to revise, and which version to deliver. The submission history and final artifact reveal which discrepancies they correct and whether those corrections survive subsequent revisions.
\item All 17 models score higher on average at final delivery than on first submission, yet the gains vary substantially with what the agent must infer. Across models, the median gain is 13.97 points for recovering a camera in a known scene and 5.10 points for reconstructing an object's geometry from reference views. Even after refinement, the highest overall score reaches only 69.10 out of 100.
\end{enumerate}
\endgroup
\clearpage
  \section{Related Work}

\noindent\textbf{Evaluating coding agents in executable environments.} SWE-bench evaluates repository-level repairs by running tests against an agent's patch~\citep{swe}. Terminal-Bench applies the same principle to realistic command-line tasks, where a verifier checks whether the agent has produced the required result~\citep{terminal}. SWE-bench Multimodal brings visual evidence into this setting through software issues whose resolution requires interpreting screenshots or other images~\citep{swemulti}. GameAtlas builds on executable verification in a setting where visual and spatial relationships are themselves part of the acceptance criteria. A submitted program may run successfully while its output places an object at the wrong depth or allows a planted foot to drift. The verifier must therefore measure properties of the artifact the code produces~\citep{design2code}.

\noindent\textbf{Visual coding and 3D creation.} Existing benchmarks examine how agents construct, edit, and refine 3D content~\citep{opcad,cadrefiner}. 3DCodeBench evaluates procedural models generated from text or image references and complements automated metrics with pairwise human comparisons~\citep{threedcode}. BlenderGym studies edits to existing Blender scenes~\citep{blendergym}. BlenderWorld evaluates editable projects created from real-world references, using evaluator agents to inspect scene structure and judge rendered views against task-specific rubrics~\citep{blenderworld}. In GameAtlas, game development experts specify both the constraint each task tests and how to measure it in the submitted artifact. Depending on the task, the evaluator can check mesh connectivity, measure how closely a camera projection matches a reference, or compare motion trajectories over time. These executable criteria score each submission directly, \textbf{without an LLM judge}.

\noindent\textbf{Game development as an evaluation setting.} GameCraft-Bench asks whether an agent can turn a specification into a playable Godot game. It evaluates the delivered project by replaying demonstrations and judging observable gameplay against a rubric~\citep{gamecraft}. GameDevBench studies a different unit of work: development tasks drawn from tutorials, in which agents modify code and visual assets. It also examines how image and video feedback affects task completion~\citep{gamedevbench}. GameAtlas takes the deliverable at an individual production stage as its unit of evaluation. Holding the scene geometry fixed, for example, makes it possible to test camera reconstruction independently of asset modeling. Scoring successive submissions under the same criteria then reveals both how closely the first implementation meets the target and which discrepancies the agent can correct within its budget.

\begin{table}[H]
\centering
\caption{Task coverage and evaluation design across benchmarks.}
\label{tab:benchmark-comparison}
\small
\renewcommand{\arraystretch}{1.24}
\setlength{\tabcolsep}{3pt}
\begin{tabular*}{\linewidth}{@{}l@{\extracolsep{\fill}}llcc@{}}
\toprule
\textbf{Benchmark} & \textbf{Coverage} & \textbf{Task unit} &
\shortstack{\textbf{Judge-free}\\\textbf{scoring}} &
\shortstack{\textbf{Hidden-test}\\\textbf{resubmission}} \\
\midrule
Terminal-Bench 2.0~\citep{terminal} & 89 terminal tasks & Terminal objective & \ding{51} & -- \\
3DCodeBench~\citep{threedcode} & 212 object classes & 3D object & \ding{51}\rlap{\textsuperscript{*}} & -- \\
BlenderWorld~\citep{blenderworld} & 9 creative domains & Blender project & -- & -- \\
GameDevBench~\citep{gamedevbench} & 4 development categories & Game feature & \ding{51} & -- \\
GameCraft-Bench~\citep{gamecraft} & 15 game families & Playable game & -- & -- \\
\midrule
\textbf{GameAtlas (ours)} & \shortstack[l]{\textbf{8 production disciplines}\\\textbf{91 tasks}} &
\shortstack[l]{\textbf{Stage-specific}\\\textbf{deliverable}} & \ding{51} & \ding{51} \\
\bottomrule
\end{tabular*}
\par\smallskip
\begin{minipage}{\linewidth}
\footnotesize
\emph{Judge-free scoring} means the delivered artifact can be scored without an LLM judge.
\emph{Hidden-test resubmission} lets an agent revise and resubmit after feedback from the hidden verifier within the same task.
\textsuperscript{*}3DCodeBench reports both automated metrics and human preferences.
\end{minipage}
\end{table}
  \section{The GameAtlas Benchmark}

\subsection{Problem Definition}

GameAtlas evaluates whether a coding agent can infer geometric, spatial, and temporal relationships from visual evidence and realize those relationships through code. Professional game development provides a setting in which both abilities have observable consequences. An inferred shape can be inspected as a mesh, a camera estimate can be used to project a known scene, and a reconstructed movement can be sampled over time.

We define a task as \(\tau=(I_\tau,E_\tau,D_\tau,V_\tau,B_\tau)\). The instruction \(I_\tau\) specifies the production objective and constraints. The initial environment \(E_\tau\) contains the reference materials, starting assets, and tools available to the agent. The delivery specification \(D_\tau\) identifies the files to be collected, \(V_\tau\) is the task-specific executable verifier, and \(B_\tau\) is the budget for intermediate submissions. The submitted artifact and its score are
\[
  a_\tau(w)=\mathop{\mathrm{Collect}}\nolimits(w,D_\tau),
  \qquad
  s_\tau(w)=V_\tau\!\left(a_\tau(w)\right)\in[0,100].
\]
Here, \(w\) denotes the current workspace state, and \(\mathop{\mathrm{Collect}}\nolimits(w,D_\tau)\) retrieves the files specified by the task's delivery requirements \(D_\tau\). The verifier \(V_\tau\) evaluates the resulting artifact, executing submitted code when the task requires it.

During a run, the agent may request feedback on up to \(B_\tau\) intermediate submissions. At termination, the verifier independently evaluates the files in the final workspace \(w_{\mathrm{final}}\), yielding \(s_\tau(w_{\mathrm{final}})\). The first submission and the final delivery thus capture two aspects of performance: the accuracy of the agent's initial implementation and the quality it ultimately achieves through inspection and revision.

\subsection{Task Coverage and Statistics}

GameAtlas comprises \textbf{91 tasks across eight disciplines} of professional game development (Table~\ref{tab:gameatlas-coverage}). These disciplines place different demands on the same underlying spatial knowledge. Image-to-mesh reconstruction, for example, requires an agent to infer a 3D surface from its 2D projections~\citep{creft}. UV unwrapping starts from a known 3D surface and asks the agent to map it into a 2D texture domain while controlling distortion and overlap~\citep{partuv}. Animation introduces a temporal dimension: a pose that is accurate in isolation may still occur too early in a movement or break contact with a surface between keyframes.

\begin{table}[htbp]
\centering
\caption{Task distribution and objectives in GameAtlas. The suite contains 89 single-stage tasks and two four-stage Sponza rendering tasks. Each Sponza task counts once.}
\label{tab:gameatlas-coverage}
\small
\setlength{\tabcolsep}{3pt}
\renewcommand{\arraystretch}{1.12}
\begin{tabular}{@{}>{\raggedright\arraybackslash}p{0.18\linewidth}r>{\raggedright\arraybackslash}p{0.27\linewidth}>{\raggedright\arraybackslash}p{0.43\linewidth}@{}}
\toprule
\textbf{Discipline} & \textbf{Tasks} & \textbf{Given} & \textbf{Objective} \\
\midrule
Image-to-mesh reconstruction & 26 & Multiple reference views of an object & Recover a 3D shape consistent with the reference views \\
Retopology & 10 & High-resolution mesh and triangle budget & Simplify geometry while preserving silhouettes, thin structures, and boundaries \\
UV unwrapping & 10 & Fixed mesh geometry & Construct a UV layout with low distortion and adequate island spacing \\
Character animation & 20 & Rigged character and reference motion & Reproduce articulation, timing, and task-specific contacts \\
Scene layout & 10 & Reference image, calibrated camera, and asset pool & Recover object counts, positions, and orientations \\
Camera reconstruction & 10 & Reference image and fixed 3D scene & Recover the camera pose and field of view that produce the target framing \\
Rendering & 3 & Scene assets, target views, and rendering environment & Implement a renderer that reproduces the target appearance across views \\
PSD composition & 2 & Layered artwork and target aspect ratio & Recompose the artwork while preserving readability, spatial relationships, and layer editability \\
\midrule
\textbf{Total} & \textbf{91} & & \\
\bottomrule
\end{tabular}
\end{table}

\subsection{Task Construction}

\textbf{Decomposing production workflows by capability.}
The eight task types in GameAtlas correspond to work performed in commercial game production. For each type, we identify what is available at the start of a production stage and what the next stage requires of its deliverable. This handoff determines which decisions we hold fixed and which visual or spatial relationships the agent must recover through code. For example, reconstructing an object from reference views leaves its three-dimensional shape unknown. Retopologizing that same kind of object supplies its shape but imposes a geometry budget, making the central decision which structures to preserve as the mesh is simplified~\citep{wildmesh}. Both tasks produce a mesh, yet they test different fundamentals: recovering structure from projections and preserving structure under a change of representation. This distinction guides how we decompose the workflow across all eight disciplines.

\textbf{Turning production briefs into verifiable tasks.}
A production brief specifies what an asset must accomplish while leaving many implementation decisions to its author. Across all eight disciplines, GameAtlas uses acceptance criteria that domain experts with extensive game studio experience developed for their own production workflows. We translate these criteria into executable checks on the artifact submitted for each task. In the shield-raising task, reaching the final pose is only part of the requirement. The arm and shield must also follow the reference motion with the intended timing. The agent reconstructs the action on a supplied rig from two synchronized views, and the verifier compares the delivered motion with the hidden 3D trajectories at corresponding instants.

\begin{figure}[H]
\centering
\includegraphics[width=\linewidth]{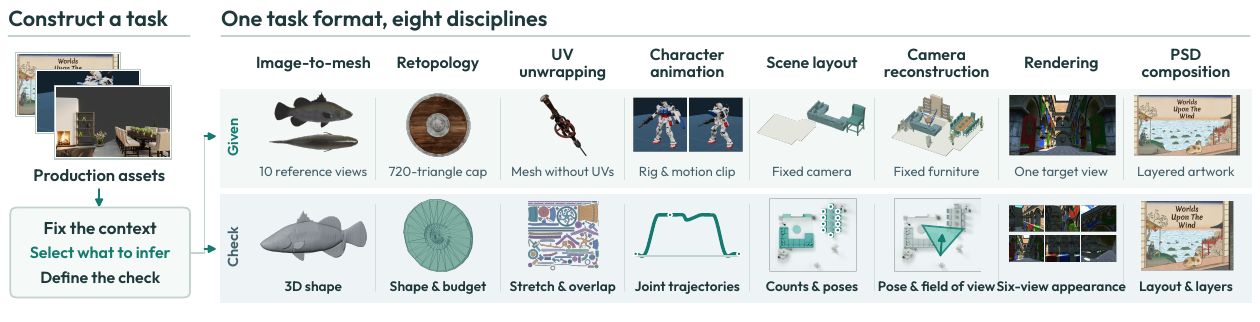}
\caption{Task construction across eight disciplines. Each column pairs the starting conditions of one task with the properties checked in the delivered artifact.}
\label{fig:task-construction}
\end{figure}

\textbf{Designing for autonomous refinement.}
We design each task so that the artifact an agent produces can inform its next revision. By inspecting its work against the available references, the agent can identify where its implementation falls short and decide which changes are warranted. It also decides when to use its limited submissions to seek feedback from the hidden verifier. Comparing the first submission with the final delivery allows us to assess whether the agent uses this evidence to reproduce the required visual and spatial relationships more precisely.
  \FloatBarrier\section{Evaluation Protocol}

GameAtlas evaluates visual coding and spatial reasoning by testing whether the relationships required by a task hold in the artifact an agent produces. We evaluate each run at two levels. For an individual delivery, we measure how closely the artifact satisfies its visual and spatial constraints under specified viewing conditions or at specified points in time. Across a run, we measure how much the agent improves its delivery after inspecting its work, obtaining feedback, and revising its code.

\subsection{Executable Artifact Evaluation}

For each task \(\tau\), the verifier \(V_\tau\) examines the form of evidence that can establish whether a requirement has been met. Lighting is observable in images rendered from specified viewpoints. Motion timing requires measurements at corresponding instants, and sustained contact requires measurements over an interval. Game development experts specify these conditions, the quantities to measure, and the tolerances for acceptable deviations. For a given evaluation target, all agents are assessed using the same executable criteria, without an LLM judge.

\begin{figure}[!htb]
\centering
\includegraphics[width=\linewidth]{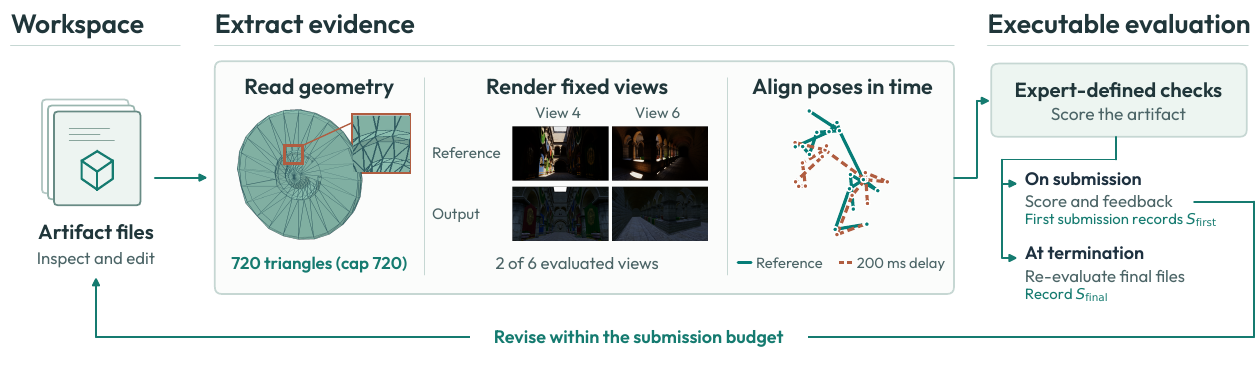}
\caption{GameAtlas evaluates delivered artifacts through task-specific checks on geometry, rendered views, or motion over time. Agents can revise their work after submission feedback. At termination, the verifier independently scores the final workspace files.}
\label{fig:evaluation-protocol}
\end{figure}

\textbf{Visual consistency across viewpoints.}
In the Sponza rendering tasks, the verifier runs the submitted renderer from six prescribed camera views and compares each output with its corresponding reference. The views expose different parts of the lighting solution: an implementation may reproduce a sunlit courtyard wall while losing detail in the shadows beneath an arcade. Each stage therefore combines the weighted mean across all six views (60\%), the mean of the two lowest-scoring views (25\%), and the lowest individual view score (15\%). This aggregation rewards agreement across the scene while retaining sensitivity to failures visible from only one or two viewpoints.

\textbf{Spatial accuracy over time.}
The spatial accuracy of an animation depends on when each pose is reached as well as where each joint is placed. To preserve this temporal constraint, the verifier samples the submitted motion at 60~Hz and compares world-space bone positions at corresponding timestamps on the reference timeline. A calibration experiment on a 15-bone clip illustrates the effect. The reference motion scores 100 and a static pose scores 5. Delaying the reference motion by 200~ms preserves the sequence of poses yet lowers the score to 62.97, exposing a timing error that an inspection of the key poses alone could miss.

\subsection{Budgeted Autonomous Iteration}

GameAtlas gives the agent control over how it refines each deliverable. It can inspect its artifact against the supplied references and revise its implementation locally without using a submission attempt. The agent decides when its own observations are sufficient to guide the next revision and when to spend a submission on feedback from the hidden verifier.

Each accepted submission uses one attempt and returns task-specific feedback along with the number of attempts remaining. Each of the 89 single-stage tasks allows up to five submissions. In each of the two Sponza rendering tasks, the agent shares a budget of six submissions across all four stages. At the end of a run, the verifier independently scores the files the agent leaves in the workspace. The submission history records the quality the agent attained along the way, while the final score reflects the improvements it retained in the delivered artifact.

\begin{figure}[!htb]
\centering
\includegraphics[width=\linewidth]{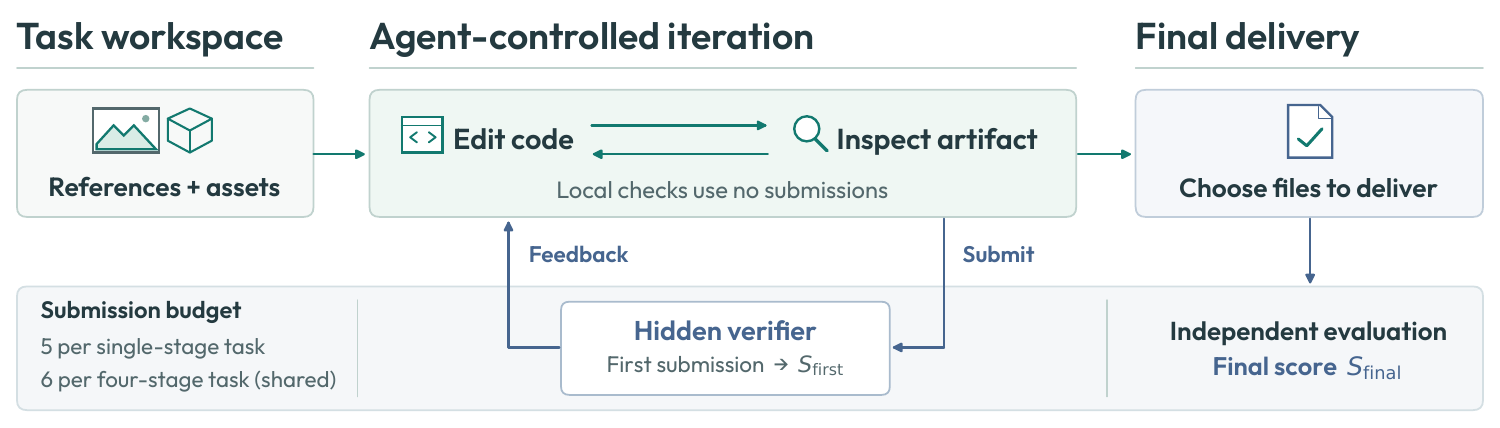}
\caption{Budgeted autonomous iteration in GameAtlas. Agents can inspect and revise artifacts locally between submissions. Each accepted submission spends one attempt on hidden-test feedback. The final score comes from an independent evaluation of the files the agent delivers.}
\label{fig:autoresearch-setting}
\end{figure}

\subsection{Metrics and Aggregation}

For each task run, we report the \textbf{first-submission score} \(S_{\mathrm{first}}\) and the \textbf{final-delivery score} \(S_{\mathrm{final}}\), both on a 0--100 scale. The first is the score of the first artifact the agent submits for evaluation. It measures what the agent achieves with local inspection before receiving feedback from the hidden tests. The final score comes from the independent evaluation at termination. If a run produces a valid final delivery without an intermediate submission, that delivery supplies both scores. The difference \(\Delta=S_{\mathrm{final}}-S_{\mathrm{first}}\) measures the net change in delivered quality and can be negative.

We aggregate first and final scores separately using the same weights. Let \(S_{\tau,r}^{(m)}\) denote the relevant score for model \(m\) on task \(\tau\) in run \(r\), and let \(R_{m,\tau}\) contain the runs included for that model and task. The overall score is
\[
S^{(m)} =
\frac{1}{\sum_{\tau=1}^{91}w_\tau}
\sum_{\tau=1}^{91} w_\tau
\left(\frac{1}{|R_{m,\tau}|}\sum_{r\in R_{m,\tau}} S_{\tau,r}^{(m)}\right).
\]
Each of the 89 single-stage tasks has weight 1, and each of the two four-stage Sponza tasks has weight 4, giving a total weight of 97. We also report an unweighted mean over the 91 tasks and scores by discipline. The unweighted mean gives each task equal influence, while the discipline scores show where models with similar overall performance differ in the visual and spatial requirements they can satisfy.
  \section{Experiments and Results}
\subsection{Experimental Setup}

All 17 models receive the same starting assets, tools, and submission budget on each of the 91 tasks. Agents work under task-specific wall-clock limits, typically four hours, without direct internet access. A separate evaluator scores submitted artifacts against private references and returns task-specific feedback that agents can use for revision (Appendix~\ref{app:execution-environment}). Our protocol calls for averaging scores across 3 runs per model. We report the mean weighted score at first submission and final delivery (Section~4.3).

\subsection{Overall Performance}

Gemini-4-Argon achieves the highest mean weighted score on first submission (60.25), leading GPT-6-Astra (55.95) by 4.30 points and GPT-6.1-Sol (54.48) by 5.77 points (Figure~\ref{fig:overall-results}). The ordering changes over the course of revision. Astra gains 13.15 points to finish first at 69.10, and Sol gains 13.69 points to finish second at 68.17. Argon also improves, reaching 63.87, but its smaller gain of 3.62 points leaves it third at final delivery. Astra thus moves from 4.30 points behind the initial leader to 5.23 points ahead of it.

\begin{figure}[H]
\centering
\includegraphics[width=\linewidth]{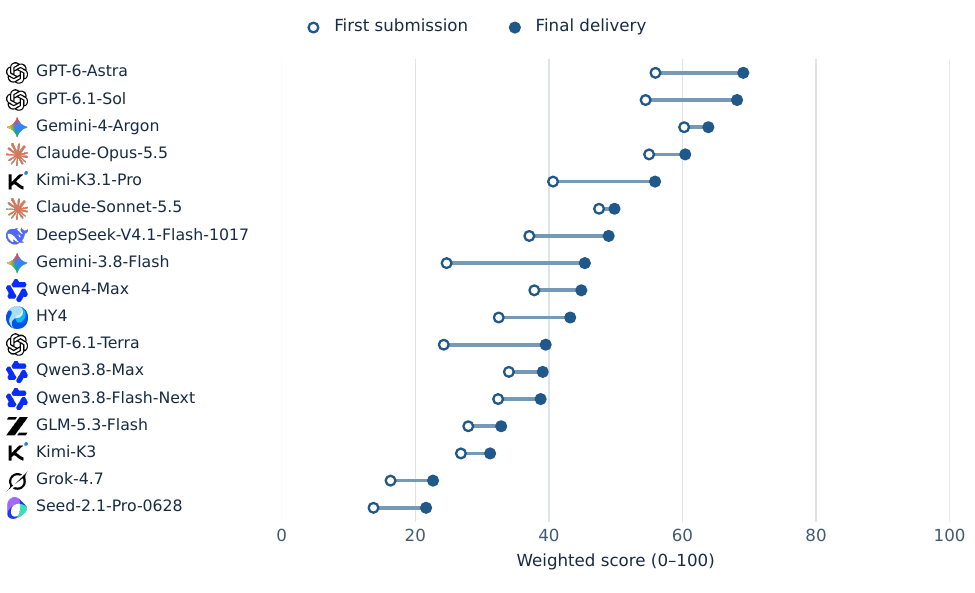}
\caption{Mean weighted scores at first submission and final delivery across 91 tasks, with models ordered by final score.}
\label{fig:overall-results}
\end{figure}

All 17 models achieve a higher mean score at final delivery than on first submission, with a median gain of 7.04 points. The magnitude of improvement varies considerably across models. Gemini-3.8-Flash records the largest gain, rising from 24.68 to 45.38 and moving from fourteenth to eighth in the overall ranking. Claude-Sonnet-5.5 starts higher at 47.51 and gains 2.32 points, finishing 4.44 points ahead of Flash. These gains reflect the full period of revision, during which agents can inspect their artifacts, continue working locally, and act on hidden-test feedback. Even with these opportunities, the highest final score across the 91 tasks is 69.10 out of 100.

\subsection{Performance Across Disciplines}

The leading models recover spatial configurations with high accuracy when the scene geometry or object assets are supplied (Figure~\ref{fig:discipline-results}). Gemini-4-Argon achieves a mean final score of 99.14 on camera reconstruction, where the agent must infer the viewpoint of a fixed scene. GPT-6-Astra reaches 94.61 on scene layout, where the camera is given and the agent must place the objects. Reconstructing the geometry itself from reference views remains considerably more difficult. Astra leads image-to-mesh reconstruction with 53.80, well below its own camera reconstruction score of 97.37. Extending reconstruction across time also remains challenging, with the highest mean animation score reaching 64.19.

\begin{figure}[!htb]
\centering
\includegraphics[width=\linewidth]{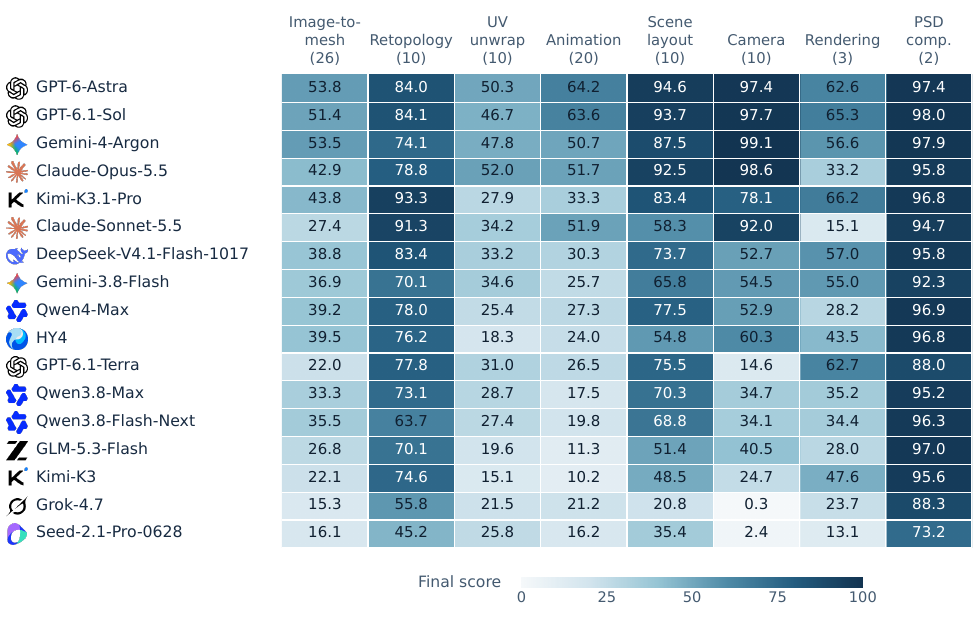}
\caption{Mean final score by model and discipline. Column headings give task counts. Every cell uses the same 0--100 color scale and reports the score to one decimal place.}
\label{fig:discipline-results}
\end{figure}

Even when the 3D geometry is supplied, performance depends on what the agent must do with the surface. Across the 17 models, the median final score is 76.18 for retopology and 28.66 for UV unwrapping. Every model scores higher on retopology. Both disciplines start from a mesh, but they place different demands on spatial reasoning. In retopology, the agent must preserve the object's shape while reducing its triangle count. In UV unwrapping, it must decide where to cut the surface and how to flatten the resulting patches into texture space while controlling distortion. This change of representation remains difficult even when the object's shape is fully specified.
  \section{Analysis of Agent Behavior}

Section~5 reports the quality of the artifacts agents deliver. We draw on tool traces, submission histories, and the artifacts themselves to examine how agents develop an initial solution, how they use visual evidence to revise it, and which errors persist at delivery.

\begin{samepage}
\subsection{Tool Use Across Agents}
\enlargethispage{0.8\baselineskip}

The first submission marks the point at which an agent can obtain feedback from the hidden verifier. Until then, it must assess its implementation using the supplied references and its own checks. We compare tool use through the first submission with tool use over the full task across all trajectories for which both endpoints can be traced. These span all 17 agents and eight disciplines. Each assistant turn that invokes tools counts as one round, even when several calls run in parallel. Table~\ref{tab:tool-use} reports per-agent medians. We take the median across agents to give each equal weight.\par
\end{samepage}

By the first submission, agents had issued a median of 74 tool calls over 66.5 rounds. Their tools had already returned 17.5 images across 12 distinct rounds, giving agents repeated opportunities to consult visual evidence while developing the initial implementation. By task completion, the corresponding medians reached 134.5 calls over 122 rounds, with 27 images returned across 17 rounds. A substantial part of the long-horizon work in GameAtlas thus takes place before the verifier assigns its first score.

The median share of tool calls devoted to shell commands and file reads was 91.7\%. Within this shared set of tools, the frequency of visual inspection varied sharply by discipline. Among the 16 agents with traces in both scene layout and UV unwrapping, the two disciplines took about 145 and 132 tool-use rounds per task, respectively. Agents received images in 47 rounds per task in scene layout and about six in UV unwrapping. Every agent in the comparison consulted visual evidence on more rounds in scene layout, even though the two workflows were comparable in overall length. Table~\ref{tab:token-cost} reports mean token usage and estimated API cost per 91-task run.

\subsection{How Do Agents Turn Visual Evidence into a Correction?}
\label{sec:visual-correction}

Gemini-4-Argon matched landmarks in the reference image to points in the supplied 3D scene. Reprojecting those points let it measure how each camera adjustment affected alignment. The penultimate submission already contained the recovered pose and field of view. Re-exporting that fit with the required metadata yielded a score of 99.72.

In Kimi-K3.1-Pro's barramundi reconstruction, the side profile was already recognizable, but broad ridges distorted the flank. Smoothing and remeshing reduced the ridges while preserving the outline. The score rose from 49.30 on the first submission to 65.04 at delivery.

For the B2 robot, GPT-6-Astra fitted the walking cycle to tracked landmarks, then compared poses at matching points in the stride to refine the swinging limbs. The score rose from 61.49 to 87.92. Across the three tasks, the agents could target a correction once they located the discrepancy in the projected image, on the mesh surface, or at a particular instant in the motion.

\subsection{Where Does Iteration Help?}

All 17 models improved on average in both camera reconstruction and image-to-mesh. Across models, the median discipline-level gain was 13.97 points for camera reconstruction and 5.10 for image-to-mesh (Fig.~\ref{fig:iteration-gains}a). Figure~\ref{fig:iteration-gains}b compares runs whose first submissions scored at least 25 but less than 50 points. By final delivery, 65.0\% of camera reconstruction runs had improved by more than 20 points, compared with 4.6\% of image-to-mesh runs. The cases in Section~\ref{sec:visual-correction} suggest one reason for this difference. Adjusting a camera can bring several landmarks into alignment at once. In mesh reconstruction, smoothing the flank of a fish can improve that surface while leaving the head, fins, and overall volume to be resolved separately.

\begin{figure}[!htb]
\centering
\includegraphics[width=\linewidth]{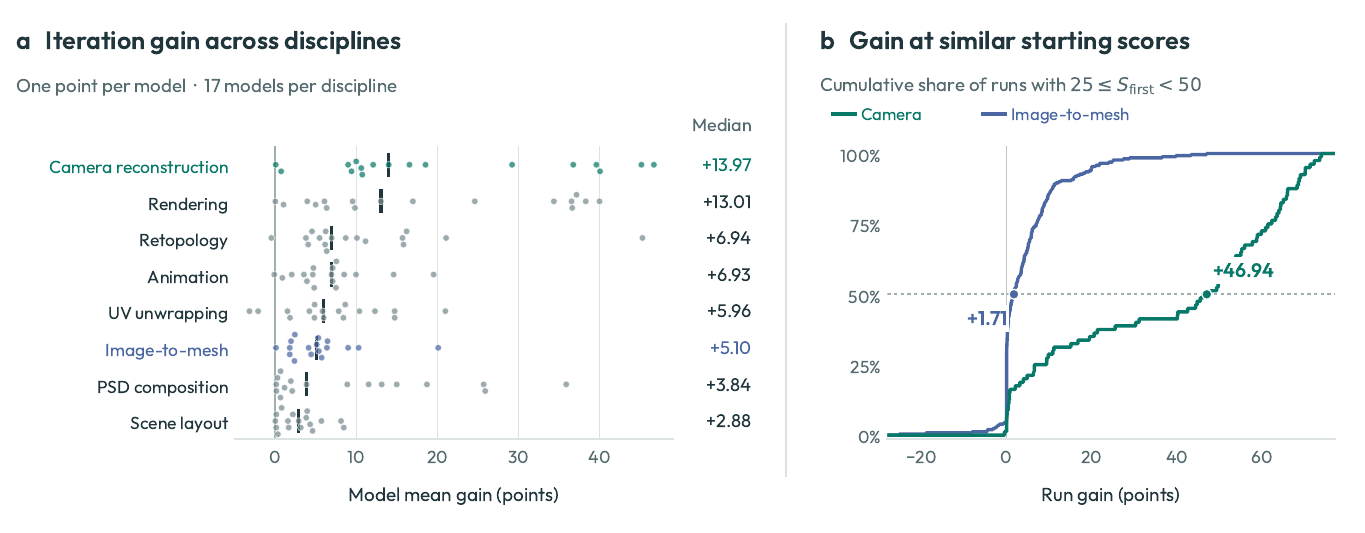}
\caption{Gain from first submission to final delivery. (a) Each dot is one model's mean gain in a discipline. Vertical bars mark medians across the 17 models. (b) For runs with \(25 \leq S_{\mathrm{first}} < 50\), each curve gives the fraction whose gain from first submission to final delivery is at most \(x\) points. Dots mark the median gain.}
\label{fig:iteration-gains}
\end{figure}

An agent can also benefit from iteration by preserving a strong solution while exploring alternatives. GameAtlas records both the scored submissions and the file ultimately delivered, allowing us to examine what happens when a later attempt falls short of an earlier one. Across 418 five-submission runs from GPT-6-Astra, Kimi-K3.1-Pro, and DeepSeek-V4.1-Flash-1017, the fifth submission finished more than one point below an earlier peak in 80 cases. \textbf{In 79 of those 80 cases, the delivered artifact scored within 0.01 points of the earlier peak or higher.} A separate GPT-6.1-Sol trace documents one way an agent preserved an earlier gain. On a cabinet retopology task, Sol saved its third mesh, which scored 89.3, before experimenting further with the drawers and back panel. After its fifth mesh scored 80.8, Sol restored the saved version, checked the export, and delivered it. Evaluating the final file captures a consequential part of autonomous refinement: the agent can try a change, assess the outcome, and decide which version is worth keeping.

\subsection{Failure Analysis}

An artifact can reproduce a recognizable feature of the reference while missing the spatial relationship that gives it its form or function. Figure~\ref{fig:supp-failure-cases} makes these discrepancies visible in the relief of the armor plates, the orientation of the armchairs, and the distance the striking arm travels.

\textbf{Recovering the outline while flattening the surface.}
In the Iron Man reconstruction task, three agents produced meshes with a recognizable character outline but lost much of the depth variation across the armor. DeepSeek-V4.1-Flash-1017, Qwen4-Max, and Kimi-K3.1-Pro scored between 88.1 and 95.5 for silhouette agreement, yet only 21.3 to 23.9 for surface relief. All three meshes also contained an excess of faces sharing the same surface orientations relative to the reference. The agents reproduced where the character ended against the background more accurately than how its surfaces projected and receded \emph{within} that boundary. Refining the outer contour alone would leave this error largely intact.

\textbf{Placing the assets but missing their orientations.}
In one furnished-room reconstruction, DeepSeek produced a physically plausible arrangement that scored 98.2 for object integrity and 97.6 on physical checks. Its object-center score was 68.7, and its rotation score fell to 47.5. The agent could assemble the supplied assets into a coherent room, but several placements failed to reproduce the reference arrangement. This separates two spatial judgments that can look similar in a finished scene: finding a plausible location for a piece of furniture and recovering the particular position and facing direction shown in the reference.

\textbf{Producing too little motion to realize the action.}
On the rocket-punch task, both DeepSeek and HY4 delivered animations in which the character moved only slightly. The root-mean-square change in joint position between successive samples was less than one tenth of the corresponding value in the reference motion. The same failure appeared in DeepSeek's wing-deployment animation and HY4's sword-stowing animation. In these cases, the main error was already present before fine timing or contact accuracy became decisive: the articulated parts did not travel far enough to perform the requested action.
  \section{Future Work}

GameAtlas currently measures visual coding and spatial reasoning through tasks built around separate assets and references. We plan to connect these tasks within a shared game project that progresses from asset creation through scene assembly and character interaction to a complete, playable game. Each stage would build on the agent's earlier deliverables while retaining its own executable checks. This would let us evaluate visual coding and spatial reasoning over longer horizons by tracing how accurately agents infer the relationships that a scene or interaction requires, preserve those relationships in code as the project grows in complexity, and correct errors that emerge only when independently developed components must work together.
  \section*{Authors}
  {\raggedright\noindent
  Yusong~Sun\textsuperscript{*}, Jingtan~Chen\textsuperscript{*}, Zixuan~Zhang\textsuperscript{*}, Jay~Li, Haoyi~Wu, Yang~Xiao,
  Wenyang~Hui, Shujun~Liu, Lichi~Zhang, Chengyue~Jiang\textsuperscript{\textdagger}, Chuan~Mu\textsuperscript{\textdaggerdbl}\par}
  {\footnotesize\noindent
  \textsuperscript{*}~Equal contribution\qquad
  \textsuperscript{\textdagger}~Project Lead\qquad
  \textsuperscript{\textdaggerdbl}~Corresponding Author\par}
  \bibliographystyle{styles/colm2026_conference}\bibliography{references}%
  \newcommand{\scoreRows}{%
GPT-6-Astra & 55.95 $\pm$ 0.43 & 69.10 $\pm$ 1.33 & 58.09 $\pm$ 0.79 & 69.69 $\pm$ 0.83 \\
GPT-6.1-Sol & 54.48 $\pm$ 1.49 & 68.17 $\pm$ 0.59 & 56.31 $\pm$ 1.48 & 68.49 $\pm$ 0.89 \\
Gemini-4-Argon & 60.25 $\pm$ 0.76 & 63.87 $\pm$ 1.71 & 60.69 $\pm$ 0.92 & 64.56 $\pm$ 1.93 \\
Claude-Opus-5.5 & 55.00 $\pm$ 2.37 & 60.41 $\pm$ 1.61 & 57.27 $\pm$ 1.79 & 62.60 $\pm$ 0.99 \\
Kimi-K3.1-Pro & 40.62 $\pm$ 2.85 & 55.89 $\pm$ 1.23 & 41.52 $\pm$ 2.99 & 55.39 $\pm$ 1.33 \\
Claude-Sonnet-5.5 & 47.51 $\pm$ 2.24 & 49.82 $\pm$ 1.41 & 50.19 $\pm$ 2.39 & 52.66 $\pm$ 1.51 \\
DeepSeek-V4.1-Flash-1017 & 37.06 $\pm$ 3.04 & 48.95 $\pm$ 1.04 & 38.49 $\pm$ 3.05 & 48.61 $\pm$ 0.99 \\
Gemini-3.8-Flash & 24.68 $\pm$ 0.12 & 45.38 $\pm$ 2.07 & 25.15 $\pm$ 0.66 & 44.85 $\pm$ 1.44 \\
Qwen4-Max & 37.81 $\pm$ 1.75 & 44.85 $\pm$ 0.52 & 39.30 $\pm$ 1.18 & 46.39 $\pm$ 0.69 \\
HY4 & 32.48 $\pm$ 0.03 & 43.19 $\pm$ 0.95 & 33.04 $\pm$ 0.11 & 43.56 $\pm$ 0.50 \\
GPT-6.1-Terra & 24.26 $\pm$ 1.58 & 39.52 $\pm$ 0.82 & 24.53 $\pm$ 1.61 & 38.15 $\pm$ 0.92 \\
Qwen3.8-Max & 34.00 $\pm$ 2.32 & 39.08 $\pm$ 2.89 & 34.48 $\pm$ 1.33 & 39.71 $\pm$ 1.62 \\
Qwen3.8-Flash-Next & 32.39 $\pm$ 0.83 & 38.76 $\pm$ 1.79 & 33.32 $\pm$ 0.16 & 39.53 $\pm$ 0.80 \\
GLM-5.3-Flash & 27.92 $\pm$ 1.95 & 32.85 $\pm$ 0.24 & 28.43 $\pm$ 0.75 & 33.68 $\pm$ 1.59 \\
Kimi-K3 & 26.82 $\pm$ 1.01 & 31.19 $\pm$ 3.54 & 27.41 $\pm$ 1.63 & 30.40 $\pm$ 3.64 \\
Grok-4.7 & 16.29 $\pm$ 1.50 & 22.65 $\pm$ 2.20 & 16.43 $\pm$ 1.57 & 23.05 $\pm$ 2.33 \\
Seed-2.1-Pro-0628 & 13.73 $\pm$ 0.72 & 21.61 $\pm$ 0.29 & 14.15 $\pm$ 0.48 & 22.82 $\pm$ 0.56 \\%
}

\section*{Supplementary Results}
\begin{table}[H]
\centering
\caption{Overall performance across 91 tasks (leaderboard score \(\pm\) sample SD).}
\label{tab:supp-results}
\small
\setlength{\tabcolsep}{4pt}
\renewcommand{\arraystretch}{1.32}
\begin{tabular*}{\linewidth}{@{}l@{\extracolsep{\fill}}rrrr@{}}
\toprule
 & \multicolumn{2}{c}{Weighted} & \multicolumn{2}{c}{Unweighted} \\
\cmidrule(lr){2-3}\cmidrule(l){4-5}
Model & First & Final & First & Final \\
\midrule
\scoreRows
\bottomrule
\end{tabular*}
\end{table}

\begin{table}[H]
\centering
\caption{Tool use by agent. Each cell is the median across task runs for the indicated agent.}
\label{tab:tool-use}
\footnotesize
\setlength{\tabcolsep}{1.6pt}
\renewcommand{\arraystretch}{1.16}
\begin{tabular*}{\linewidth}{@{}l@{\extracolsep{\fill}}*{11}{r}@{}}
\toprule
 & \multicolumn{3}{c}{Through first submission} & \multicolumn{3}{c}{Full task} & \multicolumn{5}{c}{Calls by tool over the full task} \\
\cmidrule(lr){2-4}\cmidrule(lr){5-7}\cmidrule(l){8-12}
Agent & \shortstack[r]{Tool\\rounds} & \shortstack[r]{Tool\\calls} & \shortstack[r]{Images\\returned} & \shortstack[r]{Tool\\rounds} & \shortstack[r]{Tool\\calls} & \shortstack[r]{Images\\returned} & bash & read & write & edit & Other \\
\midrule
GPT-6-Astra & 38 & 61 & 21 & 82 & 115 & 34 & 43.5 & 49.5 & 9 & 8 & 0 \\
GPT-6.1-Sol & 42 & 63 & 24 & 88 & 124 & 37 & 46 & 52 & 11 & 8 & 0 \\
Gemini-4-Argon & 58.5 & 74 & 17.5 & 122 & 148 & 27 & 104 & 33.5 & 0 & 0 & 0 \\
Claude-Opus-5.5 & 92 & 109 & 29 & 162 & 190 & 43 & 132 & 49 & 3 & 0 & 0 \\
Kimi-K3.1-Pro & 33 & 50 & 15 & 84 & 112 & 31 & 51 & 38 & 9 & 10 & 0 \\
Claude-Sonnet-5.5 & 93.5 & 105.5 & 25.5 & 160 & 167.5 & 36.5 & 139.5 & 39 & 1 & 0 & 0 \\
DeepSeek-V4.1-Flash-1017 & 88 & 104 & 30 & 185 & 210 & 46 & 157 & 46 & 0 & 0 & 0 \\
Gemini-3.8-Flash & 94 & 94 & 11 & 217 & 217 & 25 & 181.5 & 28.5 & 2 & 0 & 0 \\
Qwen4-Max & 60 & 70 & 17 & 96 & 106 & 21 & 69 & 21 & 3 & 4 & 0 \\
HY4 & 66.5 & 69.5 & 16 & 132 & 134 & 25 & 95.5 & 27 & 2 & 4 & 0 \\
GPT-6.1-Terra & 12 & 15 & 6 & 28.5 & 33 & 7 & 17.5 & 9 & 1.5 & 0 & 0 \\
Qwen3.8-Max & 68 & 74 & 11 & 102 & 108 & 15 & 84 & 15 & 5 & 3 & 0 \\
Qwen3.8-Flash-Next & 166 & 168 & 15 & 277 & 279 & 18 & 206 & 20 & 32 & 10 & 0 \\
GLM-5.3-Flash & 101 & 110 & 21 & 146 & 156 & 27 & 113 & 28 & 7 & 5 & 0 \\
Kimi-K3 & 62 & 71.5 & 19.5 & 115.5 & 134.5 & 32 & 80.5 & 35.5 & 5 & 9 & 0 \\
Grok-4.7 & 86.5 & 128.5 & 57 & 137.5 & 188 & 74 & 96 & 77 & 0 & 0 & 0 \\
Seed-2.1-Pro-0628 & 29 & 34 & 10 & 97 & 103 & 24 & 52 & 26 & 13 & 5 & 0 \\
\bottomrule
\end{tabular*}
\end{table}

\begin{table}[H]
\centering
\caption{Average token usage and estimated API cost per 91-task GameAtlas run.}
\label{tab:token-cost}
\normalsize
\setlength{\tabcolsep}{3pt}
\renewcommand{\arraystretch}{1.50}
\begin{tabular*}{\linewidth}{@{}l@{\extracolsep{\fill}}rrrr@{}}
\toprule
 & \multicolumn{3}{c}{Tokens (millions)} & \\
\cmidrule(lr){2-4}
Model & Input & Output & Total & Est. cost (USD) \\
\midrule
GPT-6-Astra & 755.38 & 6.72 & 762.10 & 1,521.94 \\
GPT-6.1-Sol & 897.18 & 7.74 & 904.91 & 315.67 \\
Gemini-4-Argon & 1,576.83 & 26.51 & 1,603.34 & 711.03 \\
Claude-Opus-5.5 & 1,613.22 & 63.95 & 1,677.16 & 2,291.05 \\
Kimi-K3.1-Pro & 1,153.51 & 9.16 & 1,162.67 & 416.88 \\
Claude-Sonnet-5.5 & 1,643.04 & 74.75 & 1,717.79 & 1,807.41 \\
DeepSeek-V4.1-Flash-1017 & 2,167.64 & 23.00 & 2,190.64 & 45.77 \\
Gemini-3.8-Flash & 2,190.19 & 16.00 & 2,206.19 & 343.22 \\
Qwen4-Max & 1,031.28 & 11.48 & 1,042.76 & 156.54 \\
HY4 & 1,045.22 & 9.45 & 1,054.68 & 55.66 \\
GPT-6.1-Terra & 80.53 & 0.52 & 81.05 & 16.67 \\
Qwen3.8-Max & 1,001.99 & 12.39 & 1,014.39 & 329.97 \\
Qwen3.8-Flash-Next & 2,811.65 & 27.00 & 2,838.64 & 60.37 \\
GLM-5.3-Flash & 1,556.96 & 20.40 & 1,577.36 & 71.47 \\
Kimi-K3 & 1,167.33 & 12.71 & 1,180.03 & 693.91 \\
Grok-4.7 & 1,048.17 & 7.13 & 1,055.30 & 827.07 \\
Seed-2.1-Pro-0628 & 658.33 & 6.50 & 664.83 & 182.90 \\
\bottomrule
\end{tabular*}
\end{table}

\clearpage
\subsection*{Qualitative Failure Cases}

Figure~\ref{fig:supp-failure-cases} examines three submissions from DeepSeek-V4.1-Flash-1017. In the Iron Man reconstruction, the outer contour closely matches the reference, but horizontal bands replace the curved plates of the chest armor. The mesh scores 95.5 for silhouette agreement and 21.3 for surface relief. In Piano Lounge, the agent places both armchairs in the same corner as in the reference, but sets them parallel; their reference orientations differ by 60\(^{\circ}\). In the Titan animation, the reference hand travels 60.8\% of the character's height from its starting position at peak extension. The submitted hand stays at its starting position throughout the sampled motion.

\begingroup

\begin{figure}[H]
  \centering
  \includegraphics[width=\linewidth]{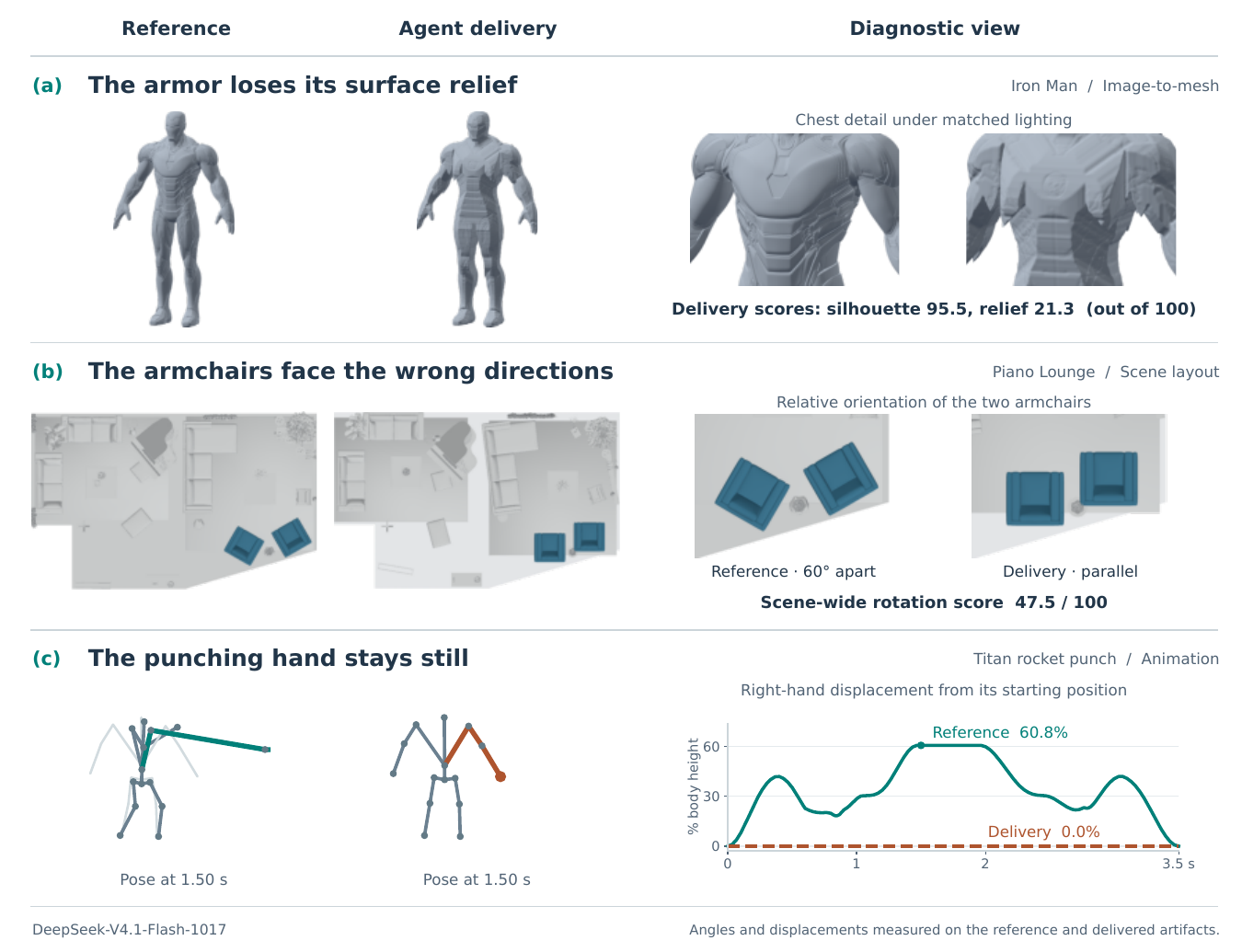}
  \caption{\textbf{Failure cases in visual coding and spatial reasoning.} Each row pairs a reference with the agent's delivery. Matched lighting reveals the lost armor relief in (a), a plan view isolates the armchair orientations in (b), and the right-hand trajectory shows the missing punch in (c). The curves give displacement from the initial hand position as a percentage of character height.}
  \label{fig:supp-failure-cases}
\end{figure}
\endgroup
  \clearpage
\appendix
\section{Inputs and Deliverables by Task Type}
\label{app:task-inventory}

\begin{table}[H]
\centering
\caption{Inputs and required deliverables for the 91 GameAtlas tasks. Counts in parentheses give the number of tasks in each discipline.}
\label{tab:task-inventory}
\setlength{\tabcolsep}{8pt}
\renewcommand{\arraystretch}{1.2}
\begin{tabular}{@{}>{\raggedright\arraybackslash}p{\dimexpr0.49\linewidth-\tabcolsep\relax}>{\raggedright\arraybackslash}p{\dimexpr0.51\linewidth-\tabcolsep\relax}@{}}
\toprule
\textbf{Given} & \textbf{Required delivery}\\
\midrule
\textbf{Image-to-mesh reconstruction (26).}\newline
Ten orthographic views of an object at a common scale, with the viewing directions specified. &
An editable 3D mesh (FBX) whose shape, thickness, and part arrangement are consistent across the supplied views.\\
\addlinespace[1em]
\textbf{Retopology (10).}\newline
A high-resolution source mesh and a task-specific triangle budget. &
A low-poly mesh (FBX) that preserves the source shape, structural features, scale, and orientation within the triangle budget.\\
\addlinespace[1em]
\textbf{UV unwrapping (10).}\newline
A fixed mesh without UV coordinates. &
The mesh (FBX) with a complete UV layout in a single \([0,1]^2\) tile. The surface geometry and normals must be preserved.\\
\addlinespace[1em]
\textbf{Character animation (20).}\newline
A rigged character, a single-view or synchronized two-view motion reference, and a specified timeline. &
An animated Blender scene that reproduces the reference poses, timing, and relevant contacts using the supplied rig.\\
\addlinespace[1em]
\textbf{Scene layout (10).}\newline
A reference image, a bare floor, a calibrated camera, and a pool of movable assets. &
A placement script that recreates the scene from the initial workspace by selecting assets, instantiating the required number of copies, and setting their transforms. The camera stays fixed.\\
\addlinespace[1em]
\textbf{Camera reconstruction (10).}\newline
A reference image and a furnished 3D scene with fixed object transforms. &
A camera specification (JSON) giving the pose and horizontal field of view that reproduce the reference framing.\\
\addlinespace[1em]
\textbf{Rendering (3).}\newline
Scene assets, a rendering workspace, and target images. The two Sponza tasks supply successive lighting targets. The chess task supplies a faulty renderer and a target view. &
Executable renderer source that produces the required appearance from the prescribed viewpoints, using the supplied scene assets.\\
\addlinespace[1em]
\textbf{PSD composition (2).}\newline
Layered \(16{:}9\) artwork and a brief specifying which subjects, text, and spatial relationships must survive adaptation to \(4{:}3\). &
An editable \(1200\times900\) PSD retaining the required named layers, together with a matching flattened PNG.\\
\bottomrule
\end{tabular}
\end{table}
\clearpage
  \section{Executable Scoring Rules and Calibration}
\label{app:verifiers}

For a task \(\tau\), the verifier extracts measurements from the delivered artifact \(a\) and computes a score \(S_\tau(a)\in[0,100]\). The measurements, tolerances, aggregation rules, and validity checks are specified before evaluation. No LLM judge is used. Below, component qualities are normalized to \([0,1]\) unless expressed in points. We write \([x]_+=\max(0,x)\) and \(\operatorname{clip}(x)=\min(1,\max(0,x))\).

\subsection{Image-to-Mesh Reconstruction}
The verifier measures four properties of the recovered geometry: surface agreement \(F\), silhouette agreement \(I\), local structure \(U\), and surface detail \(D\). Surface agreement uses distances in both directions between the submitted and reference meshes. An omitted fin therefore incurs an error from reference to submission, even if every point on the submitted fish lies close to the reference surface.

The 26 tasks use three scoring variants (Table~\ref{tab:mesh-rules}). Let \(q=(F,I,U,D)\), and let \(G(q)\) be the smoothed weighted geometric mean used by the mesh verifier,
\[
  \begin{aligned}
    G(q)&=\frac{\prod_j(0.05+0.95q_j)^{w_j}-0.05}{0.95},\\
    w&=(0.825\times0.4375,\;0.825\times0.3125,\;0.825\times0.25,\;0.175).
  \end{aligned}
\]

\begin{table}[H]
\centering
\caption{Image-to-mesh scoring rules by task group. The last column gives the final score on the unit scale.}
\label{tab:mesh-rules}
\small
\renewcommand{\arraystretch}{1.22}
\begin{tabular}{@{}lr@{\hspace{1.6em}}l@{}}
\toprule
Task group & Tasks & \(S_{\rm mesh}/100\) \\
\midrule
Numbered assets & 10 & \(\bigl[G(q)-d_{\rm geom}-d_{\rm orient}\bigr]_+\)\\
Named characters & 6 & \(\bigl[G(q)(0.4+0.6L)(1-p_{\rm geom})-d_{\rm orient}\bigr]_+\)\\
Expert assets & 10 & \(K_\tau\bigl[B(0.68+0.32L)(1-p_{\rm geom})H-d_{\rm orient}\bigr]_+\)\\
\bottomrule
\end{tabular}
\end{table}

Here \(L\) measures local depth agreement across image tiles. For the expert assets, \(B=0.32F+0.10I+0.16U+0.12D+0.30R\), where \(R\) measures agreement in surface relief. The factor \(H=\sqrt{\min(1,\max(F,I)/0.25)}\) limits credit from local detail when the overall shape is poor. The terms \(d_{\rm geom}\), \(p_{\rm geom}\), and \(d_{\rm orient}\) are geometric and orientation penalties. \(K_\tau\) collects the applicable reference-relative corrections for excess planar faceting and, on designated assets, local surface structure. These rules distinguish an accurate outline from an accurate 3D surface.

\subsection{Retopology}
Retopology measures how well a low-poly mesh preserves the source shape under a triangle budget. The verifier starts from 100 points and subtracts measured defects. Grouping the implementation's deductions by their physical meaning gives
\[
  S_{\rm retopo}=\bigl[100-D_{\rm shape}-D_{\rm topology}-D_{\rm export}-D_{\rm budget}\bigr]_+ .
\]
\(D_{\rm shape}\) includes thickness error, missing surface area, and loss of local structure. \(D_{\rm topology}\) includes poorly shaped faces, redundant edges on flat regions, and mesh-integrity defects. \(D_{\rm export}\) covers incorrect export transforms. For a submitted mesh with \(n\) triangles and a task limit \(b\), the budget deduction in points is
\[
  D_{\rm budget}=100\,\operatorname{clip}\!\left(1.5\left(\frac{n-b}{b}-0.05\right)\right).
\]
The first 5\% above the limit carries no budget deduction. Beyond that tolerance, each additional 10\% of the budget costs 15 points until the deduction saturates. Finishing below budget incurs an under-modeling deduction only to the extent that reference geometry is also missing.

\newpage
\subsection{UV Unwrapping}
The verifier first checks that the original surface and normals are preserved and that all UV faces are finite, nondegenerate, and contained in a single unit atlas. Let \(v\) equal one when these checks pass and zero otherwise. For continuity \(C\), distortion quality \(D\), texel-density consistency \(T\), atlas coverage \(A\), and chart regularity \(R\), the score is
\[
  \begin{aligned}
    S_{\rm UV}=100v\bigl[1
       &-0.60(1-C)-0.60(1-D)-0.10(1-T)\\[-1pt]
       &-0.25(1-A)-0.60(1-R^{1.25})\bigr]_+ .
  \end{aligned}
\]
The coefficients are maximum deductions and can sum to more than one. Two terms illustrate what the verifier measures. If \(u\) is atlas occupancy and \(p\) is the measured padding shortfall, then \(A=u^2\,0.25/(0.25+p)\). If \(o\) is the fraction of surface area affected by nonexempt overlap, then \(D=D_{\rm local}\,0.01/(0.01+o)\). Tightly packing the islands can improve occupancy while reducing credit if the gaps fall below the four-pixel target at \(1024^2\) resolution.

\subsection{Character Animation}
The verifier samples the candidate and reference motions at corresponding times at 60~Hz. Let \(P\) measure body-pose fidelity, \(C\) the required contacts or attachments, and \(A\) the motion of an action-specific surface such as a shield or wing. For the applicable components \(J_\tau\),
\[
  Q_\tau=\frac{\sum_{j\in J_\tau}w_jq_j}{\sum_{j\in J_\tau}w_j},
  \qquad
  T=1-\sqrt{\operatorname{mean}_{r,t}\!\left[(1-s_{r,t})^2\right]},
  \qquad
  S_{\rm anim}=100\min(Q_\tau T,c_\tau).
\]
Here \(s_{r,t}\) is temporal stability for body region \(r\) at time \(t\), and \(c_\tau\) is the task's motion ceiling. Eighteen tasks use nominal weights of 0.85 for pose and 0.15 for contacts. The shield-raising and wing-deployment tasks assign 0.40 to pose, 0.45 to the action surface, and 0.15 to contacts. The ceiling limits the credit available to a nearly static delivery. The expert-motion tasks additionally make the ceiling sensitive to the direction and amplitude of the recovered movement.

\subsection{Scene Layout}
With the camera fixed, the verifier matches submitted object instances to the reference and measures instance integrity \(I\), center position \(X\), orientation \(O\), and physical plausibility \(P\). The physical term combines support and collision quality as \(P=\sqrt{q_{\rm support}q_{\rm collision}}\). Let \(b\) be the verifier's in-bounds fraction and \(v_{\rm ctx}\) indicate that the immutable scene context is preserved. For a nonempty, valid delivery,
\[
  S_{\rm layout}=100\,v_{\rm ctx}\,\operatorname{clip}\!\left(\frac{b-0.80}{0.15}\right)
        \left(\prod_{q\in\{I,X,O,P\}}\max(q,10^{-3})\right)^{1/4}.
\]
The geometric mean makes each spatial requirement consequential. A chair can have the correct identity, center, and orientation and still lose credit if it intersects the floor. The in-bounds factor reaches full credit at 95\% and falls to zero at 80\%.

\subsection{Camera Reconstruction}
With the objects fixed, the verifier measures anchor reprojection and depth \(A\), camera translation \(X\), rotation \(O\), and focal length \(F\). Translation, rotation, and focal errors are converted to qualities using \(q(e;\sigma)=1/(1+(e/\sigma)^2)\), so the tolerance \(\sigma\) has a direct interpretation: an error of that magnitude earns half credit for the component. At each tolerance scale \(s\), the score is
\[
  Q_s=v_{\rm cam}\,A_s^{0.50}X_s^{0.20}O_s^{0.20}F_s^{0.10},
  \qquad
  S_{\rm camera}=100\,v_{\rm scene}\bigl(0.25Q_1+0.75Q_{0.075}\bigr).
\]
The precise band uses 7.5\% of the broad-band tolerances. The anchor term measures both where a 3D point projects and its depth relative to the camera. This makes the score sensitive to a camera that has been moved backward and given a different focal length even when the resulting framing is similar~\citep{wide2long}. \(v_{\rm cam}\) measures transform validity, and \(v_{\rm scene}\) checks that the fixed scene was preserved.

\newpage
\subsection{Rendering}
The Sponza tasks measure lighting across six prescribed views. Let \(x_i\) be the image score for view \(i\), \(\bar{x}_w\) the weighted mean, and \(x_{(1)}\le x_{(2)}\) the two lowest view scores. The visual score at lighting stage \(\ell\) is
\[
  V_\ell=0.60\bar{x}_w+0.25\frac{x_{(1)}+x_{(2)}}{2}+0.15x_{(1)}.
\]
The two renderers use different final-score rules. For OpenGL, let \(\ell\in\{0,1,2,3\}\) be the stage reached and \(g_{\rm src},g_{\rm renderer}\) the source and renderer checks. For Three.js, the delivered renderer is evaluated against the final indirect-lighting target:
\[
 S_{\rm OpenGL}=25\bigl(\ell+V_\ell g_{\rm src}g_{\rm renderer}\bigr),
 \qquad S_{\rm Three.js}=100V_3g_{\rm src}.
\]
The chess-scene repair task instead targets material appearance. Each of four views receives \(x_i=0.20F_i+0.25M_i+0.50H_i+0.05E_i\), measuring full-frame agreement, material-region agreement, transmission and highlights, and successful rendering. Its final score is \(100(0.50\bar{x}+0.30\bar{x}_{\rm worst\,2}+0.20x_{(1)})\). The transmission term gives the glass pieces substantial influence even when they occupy few pixels.

\subsection{Layered Image Composition}
Both banner tasks inspect the editable PSD, register its layers against the source artwork, and compare its composite with the submitted PNG. Their content-specific rules differ. For the bridge banner, let \(I,L,V,R,E\) denote asset identity, layout, visibility, compositional relationships, and export consistency. For the seaside banner, let \(I_j\) and \(L_j\) denote identity and placement for layer \(j\). On valid deliveries,
\[
  \begin{aligned}
    S_{\rm bridge}  &=\min\!\left(20I+20L+15V+25R+20E,\; C_{\rm bridge}\right),\\
    S_{\rm seaside} &=\min\!\left(\frac{1}{8}\sum_{j=1}^{8}(60I_j+40L_j),\; C_{\rm seaside}\right).
  \end{aligned}
\]
Each \(C\) is the lowest applicable content or delivery ceiling, with a default of 100. In the bridge task, the verifier measures contact between the travelers' feet and the bridge. In the seaside task, it measures support beneath the cat's lower contour and the visible fraction of individual title regions. A high average across intact layers therefore cannot erase a critical local failure.

\subsection{Calibration with Controlled Perturbations}
We isolate individual terms by changing one property of a reference while holding the others fixed. The first two rows of Table~\ref{tab:verifier-calibration} report complete task scores. The last row reports the six-view visual score \(100V_0\) for the ambient-lighting stage.

\begin{table}[H]
\centering
\caption{Verifier responses to controlled perturbations. Each row compares a reference with two changes to the same artifact. Scores are on a 0--100 scale.}
\label{tab:verifier-calibration}
\small
\renewcommand{\arraystretch}{1.16}
\setlength{\tabcolsep}{5pt}
\begin{tabular*}{\linewidth}{@{\extracolsep{\fill}}lrrr@{}}
\toprule
Verifier & Reference & Perturbation 1 & Perturbation 2 \\
\midrule
Animation & Original clip & 200\,ms delay & Static pose\\
15-bone clip & \(100.00\) & \(62.97\) & \(5.00\)\\
\addlinespace
Camera reconstruction & \(70^\circ\) field of view & \(72^\circ\) & \(75^\circ\)\\
HSSD 102344022~\citep{hssd} & \(100.00\) & \(41.90\) & \(27.43\)\\
\addlinespace
Sponza Three.js & Original views & Blur \(\sigma=1\) px & Blur \(\sigma=4\) px\\
Ambient lighting & \(100.00\) & \(91.38\) & \(78.23\)\\
\bottomrule
\end{tabular*}
\end{table}

The delayed animation preserves the pose sequence but changes its timing. The camera perturbation preserves the exact reference transform but changes the lens. The blur perturbation preserves viewpoint and illumination but removes image detail. Together, these controls show which property a score responds to and how strongly it responds in a real task.
  \clearpage
\section{Execution Environment and Evaluation Integrity}
\label{app:execution-environment}

\textbf{Wall-clock limits.}
Table~\ref{tab:agent-wall-clock} gives the time available to the agent on each task. Each Sponza task allocates time separately to its four rendering stages. The verifier has its own execution timeout.

\begin{table}[H]
\centering
\caption{Agent wall-clock limits across the 91 tasks. Stage limits are listed in execution order.}
\label{tab:agent-wall-clock}
\small
\renewcommand{\arraystretch}{1.18}
\begin{tabular*}{\linewidth}{@{}l@{\extracolsep{\fill}}rll@{}}
\toprule
Task group & Tasks & Agent wall-clock limit & Stage limits (L0--L3) \\
\midrule
Non-rendering tasks & 88 & 4 hours per task & Single stage \\
Beautiful Game render repair & 1 & 3 hours & Single stage \\
Sponza OpenGL & 1 & 12 hours & 3 hours each \\
Sponza Three.js & 1 & 4 hours & 45, 60, 90, 45 minutes \\
\bottomrule
\end{tabular*}
\end{table}

\textbf{Network access and scoring isolation.}
Agents work with the supplied assets in a sandbox without direct internet access. Dedicated interfaces provide model inference and submission feedback. Hidden tests and private references remain outside the agent's workspace. In the rendering tasks, the submitted renderer produces images in an environment without access to the private reference views. An independent scorer then compares those images with the references and, where required, inspects the submitted source. This separation allows the scorer to measure visual agreement without running agent code alongside the reference data.
}
\documentclass{article}

\makeatletter
\def\input@path{{styles/}}
\makeatother
\usepackage[preprint]{styles/colm2026_conference}
\usepackage{fontspec}

\normalfont
\usepackage{microtype}
\usepackage{graphicx}
\usepackage{trimclip}
\usepackage{xcolor}
\usepackage{booktabs}
\usepackage{array}
\usepackage{colortbl}
\usepackage{float}
\usepackage{tikz}
\usepackage{tcolorbox}
\usepackage{pgfplots}
\pgfplotsset{compat=1.18}
\usepgfplotslibrary{groupplots}
\usepackage{hyperref}
\usepackage{url}
\setlength{\headheight}{16pt}

\setlength{\textwidth}{6.5in}
\setlength{\oddsidemargin}{0in}
\setlength{\evensidemargin}{0in}

\definecolor{abyss}{HTML}{121D36}
\definecolor{polarnight}{HTML}{1A2947}
\definecolor{nebula}{HTML}{2B3F66}
\definecolor{steeltrail}{HTML}{6D87BD}
\definecolor{skytrail}{HTML}{8FA8D8}
\definecolor{starlight}{HTML}{DFE7F5}
\definecolor{warmstar}{HTML}{E8D9C4}
\definecolor{allsparkwordmark}{HTML}{16233F}
\definecolor{allsparkspark}{HTML}{4A659C}
\definecolor{electricblue}{HTML}{3866FF}
\definecolor{covercream}{HTML}{EEF3FA}
\definecolor{coveraccent}{HTML}{3866FF}
\colorlet{pevekpurple}{skytrail}
\colorlet{bargray}{steeltrail}
\colorlet{barlgray}{starlight}

\newfontfamily\outfit[
  NFSSFamily=GameAtlasOutfit,
  BoldFont=assets/fonts/Outfit-SemiBold.ttf
]{assets/fonts/Outfit-Regular.ttf}


\hypersetup{
  colorlinks=true,
  linkcolor=electricblue,
  citecolor=electricblue,
  urlcolor=coveraccent,
  filecolor=electricblue
}
\newcommand{\coverlink}[2]{%
  \href{https://#2}{%
    \makebox[12pt][c]{\raisebox{-0.27\height}{#1}}%
    \hspace{4pt}#2}}
\setcitestyle{numbers,square,comma,sort&compress}

\newcommand{\reporttitle}{Bringing LLM Agents Back to Basics\\in Visual Coding and Spatial Reasoning for Game Development}
\title{\reporttitle}
\author{AllSpark Team}

\begin{document}

\fancyhead{}
\renewcommand{\headrulewidth}{0pt}
\color{abyss}
\thispagestyle{empty}

\vspace*{-0.44in}
\begin{tcolorbox}[
  width=\linewidth,
  colback=covercream,
  colframe=covercream,
  boxrule=0pt,
  arc=14pt,
  outer arc=14pt,
  boxsep=0pt,
  left=20pt,
  right=20pt,
  top=13pt,
  bottom=11pt
]
  {\outfit\fontsize{15}{18}\selectfont\bfseries\centering
    \textcolor{coveraccent}{GameAtlas:}\hspace{0.25em}\reporttitle\par}
  \vspace{0.8em}
  {\centering\bfseries AllSpark Team\par}

  \vspace{0.75em}
  \begingroup
  \normalfont
  \setlength{\parindent}{0pt}
  \setlength{\parskip}{0.4em}
  Games are a compelling showcase for frontier coding agents. The result is immediately visible and playable, while building a complete game demands long-horizon visual coding and spatial reasoning. Existing game-generation benchmarks often focus on end-to-end performance. In a finished game, the effects of many interdependent decisions are difficult to disentangle, making it hard to measure individual capabilities consistently and objectively or pinpoint where an agent falls short. One-shot generation is also a poor proxy for professional production, where the value of a deliverable depends on whether it can be refined and used in the next stage. The distinct stages of professional game development offer a natural framework for assessing these capabilities individually. We introduce \textbf{GameAtlas}, a benchmark that breaks down eight stages of professional game development into 91 independently verifiable tasks. Each task requires an agent to satisfy specific visual or spatial constraints through code. Game development experts translate these constraints into executable criteria that score the submitted artifacts directly, without an LLM judge. Agents can inspect their artifacts, revise their code, and resubmit to hidden tests within each task's submission budget. This design lets us measure both the accuracy of an agent's initial implementation and its ability to correct errors using feedback. We evaluate 17 models on GameAtlas. GPT-6-Astra achieves the highest final score, improving from 55.95 on its first submission to 69.10 out of 100 on its last. Even with repeated opportunities to test and refine their work, frontier coding agents still struggle to satisfy the precise visual and spatial constraints of professional tasks.

  \par
  \endgroup

  \vspace{0.65em}
  \noindent
  \begin{minipage}[c]{0.76\linewidth}
    \outfit\fontsize{8.4}{10.2}\selectfont
    \renewcommand{\arraystretch}{1.28}
    \begin{tabular}{@{}l@{}}
      \coverlink{\includegraphics[height=10.5pt]{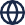}}{allspark-research.github.io/GameAtlas/} \\
      \coverlink{\includegraphics[height=10.5pt]{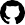}}{github.com/AllSpark-Research/GameAtlas} \\
      \coverlink{\includegraphics[height=10.5pt]{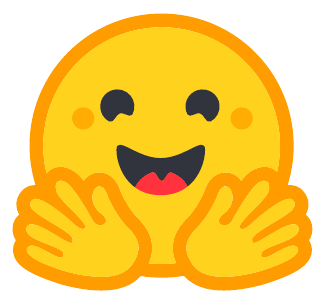}}{huggingface.co/datasets/AllSpark-Research/GameAtlas}
    \end{tabular}
  \end{minipage}%
  \hfill
  \begin{minipage}[c]{0.20\linewidth}
    \raggedleft
    \includegraphics[height=18pt]{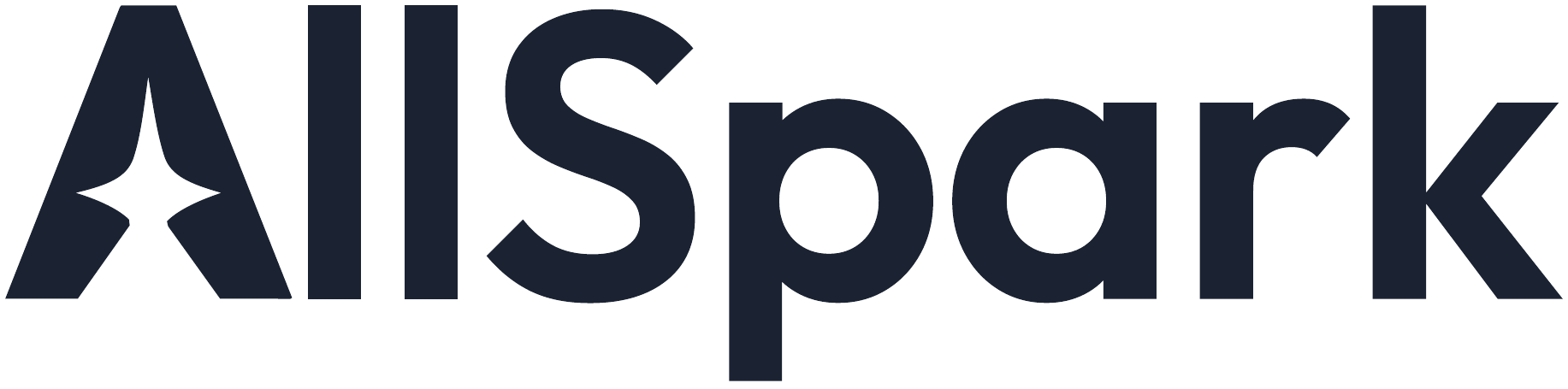}%
  \end{minipage}
\end{tcolorbox}
\end{document}